\documentclass[letterpaper, 10 pt, conference]{ieeeconf}  
\IEEEoverridecommandlockouts                              
\usepackage{hyperref}
\usepackage{cite}
\usepackage{breakcites}
\usepackage{graphicx}
\usepackage{amssymb}
\usepackage{amsmath}
\usepackage{amsfonts}
\usepackage{booktabs}
\usepackage{siunitx}
\usepackage{multirow}
\usepackage{boldline}
\usepackage{xcolor}         

\definecolor{ben}{rgb}{0.9,0.,0.5}

\definecolor{jung}{rgb}{1.0,0.7,0.0}

\definecolor{todo}{rgb}{1.0, 0.0, 0.}

\title{\LARGE \bf MonoGraspNet: 6-DoF Grasping with a Single RGB Image}

\author{Guangyao Zhai$^{1,\ast}$, Dianye Huang$^{1,\ast}$, Shun-Cheng Wu$^{1}$, HyunJun Jung$^{1}$,\\  Yan Di$^{1,\dag}$, Fabian Manhardt$^{2}$, Federico Tombari$^{1,2}$, Nassir Navab$^{1,3}$ and Benjamin Busam$^{1}$
\thanks{* Contribute equally. $\dag$ Corresponding author.}
\thanks{$^{1}$ Technical University of Munich (TUM), Munich, Germany.}
\thanks{$^{2}$ Google.}
\thanks{$^{3}$ Johns Hopkins University, Baltimore, MD, USA.}
}

\begin{document}
\maketitle
\thispagestyle{empty}
\pagestyle{empty}

\begin{abstract}
6-DoF robotic grasping is a long-lasting but unsolved problem.
Recent methods utilize strong 3D networks to extract geometric grasping representations from depth sensors, demonstrating superior accuracy on common objects but performing unsatisfactorily on photometrically challenging objects, e.g., objects in transparent or reflective materials.
The bottleneck lies in that the surface of these objects can not reflect accurate depth due to the absorption or refraction of light. 
In this paper, in contrast to exploiting the inaccurate depth data, we propose the first RGB-only 6-DoF grasping pipeline called \textit{MonoGraspNet} that utilizes stable 2D features to simultaneously handle arbitrary object grasping and overcome the problems induced by photometrically challenging objects. 
MonoGraspNet leverages a keypoint heatmap and a normal map to recover the 6-DoF grasping poses represented by our novel representation parameterized with 2D keypoints with corresponding depth, grasping direction, grasping width, and angle. 
Extensive experiments in real scenes demonstrate that our method can achieve competitive results in grasping common objects and surpass the depth-based competitor by a large margin in grasping photometrically challenging objects.
To further stimulate robotic manipulation research, we annotate and open-source a multi-view grasping dataset in the real world containing 44 sequence collections of mixed photometric complexity with nearly 20M accurate grasping labels.

\end{abstract}


\section{Introduction}
Humans visually perceive the world with passive RGB views and are able to perform sophisticated interactions with objects even if the objects are unseen previously, translucent, reflective, or transparent. 
Robotic grasping primarily relies on RGB cameras~\cite{guo2016object}, or active depth sensors such as ToF~\cite{maldonado2010robotic}, LiDAR~\cite{8593669}, and active stereo~\cite{5379597}, to perform arbitrary object grasping using the simplified grasping representation in $SE(2)$ or $SE(3)$ with no prior knowledge such as explicit object models or category information.
Existing learning–based approaches can be categorized in two directions, namely planar grasping and 6-DoF grasping.
Planar grasping~\cite{ren2015faster, he2017mask, cai2018cascade, redmon2016you} relies on a simple but effective grasping representation, which defines grasps as oriented bounding boxes. 
Although such a low DoF grasp representation reduces the task to a simple detection problem, it limits the performance in 3D manipulation tasks. %
%
%
6-DoF grasping enjoys more dexterity than planar grasping, which is suitable for handling complex scenarios. Most 6-DoF grasping methods~\cite{liang2019pointnetgpd, sundermeyer2021contact, mousavian20196, wu2020grasp, fang2020graspnet} extract geometric information from dense input and use it to score the grasp candidates. Despite the satisfactory grasping performance on common objects, even if the shape is unknown, the reliance on the dense 3D information makes them vulnerable when the input sensing is unstable, especially on
transparent or highly reflective surfaces~\cite{jung2022my}. %
Cross-modality fusion approaches for perception tasks can be used to combine RGB images with other sensor inputs~\cite{lopez2020project, gasperini2021r4dyn, verdie2022cromo}. 
This allows~\cite{sajjan2020clear,fang2022transcg} to deal with the aforementioned limitation and shows promising results. 
Nevertheless, these methods still rely on external depth information as indispensable input, making the methods less convenient and expensive. 
Motivated by this, we aim to fully explore the ability of the easily accessible and stable RGB modality on the 6-DoF robotic grasping task without any depth input.


\begin{figure}[t]
    \centering
    \includegraphics[width=0.48\textwidth, angle=0]{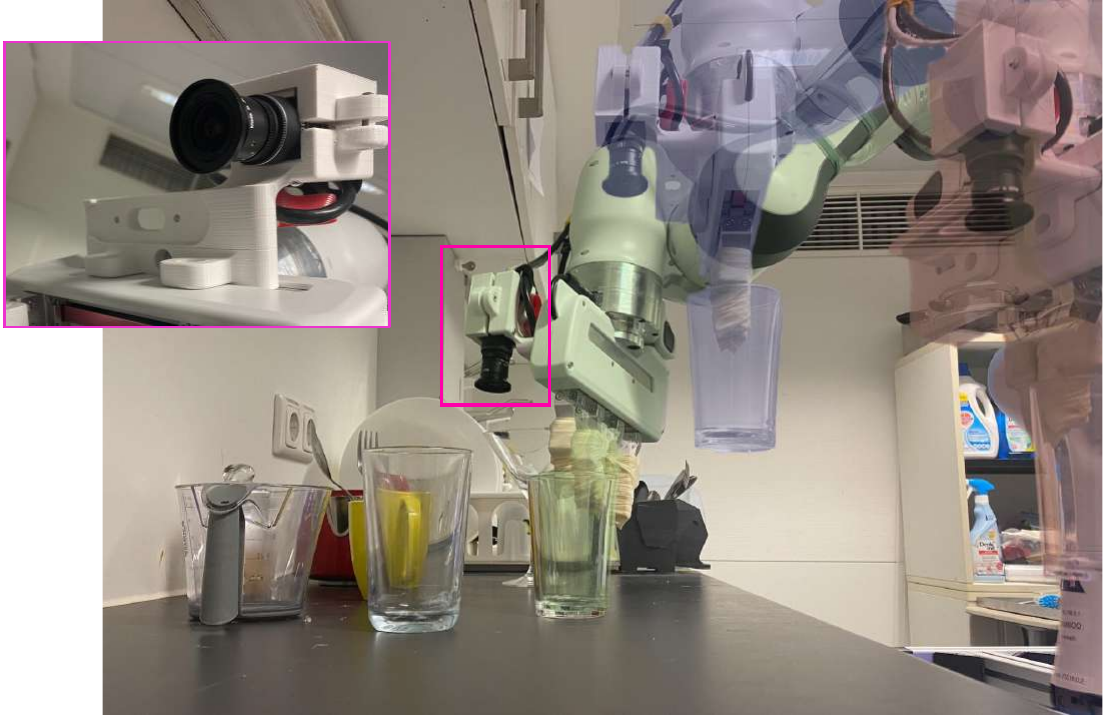}
    \caption{A Franka robot mounting an RGB camera driven by MonoGraspNet is performing transparent object grasping of a glass cup in a household scenario. The crop on the left shows the custom mount for the RGB camera on the gripper and the color overlays (green-blue-red) illustrate the grasping process for this photometrically challenging object.}
    \vspace{-3mm}
    \label{fig:cover}
\end{figure}

\begin{figure*}[t]
    \centering
    \includegraphics[width=1.0\textwidth, angle=0]{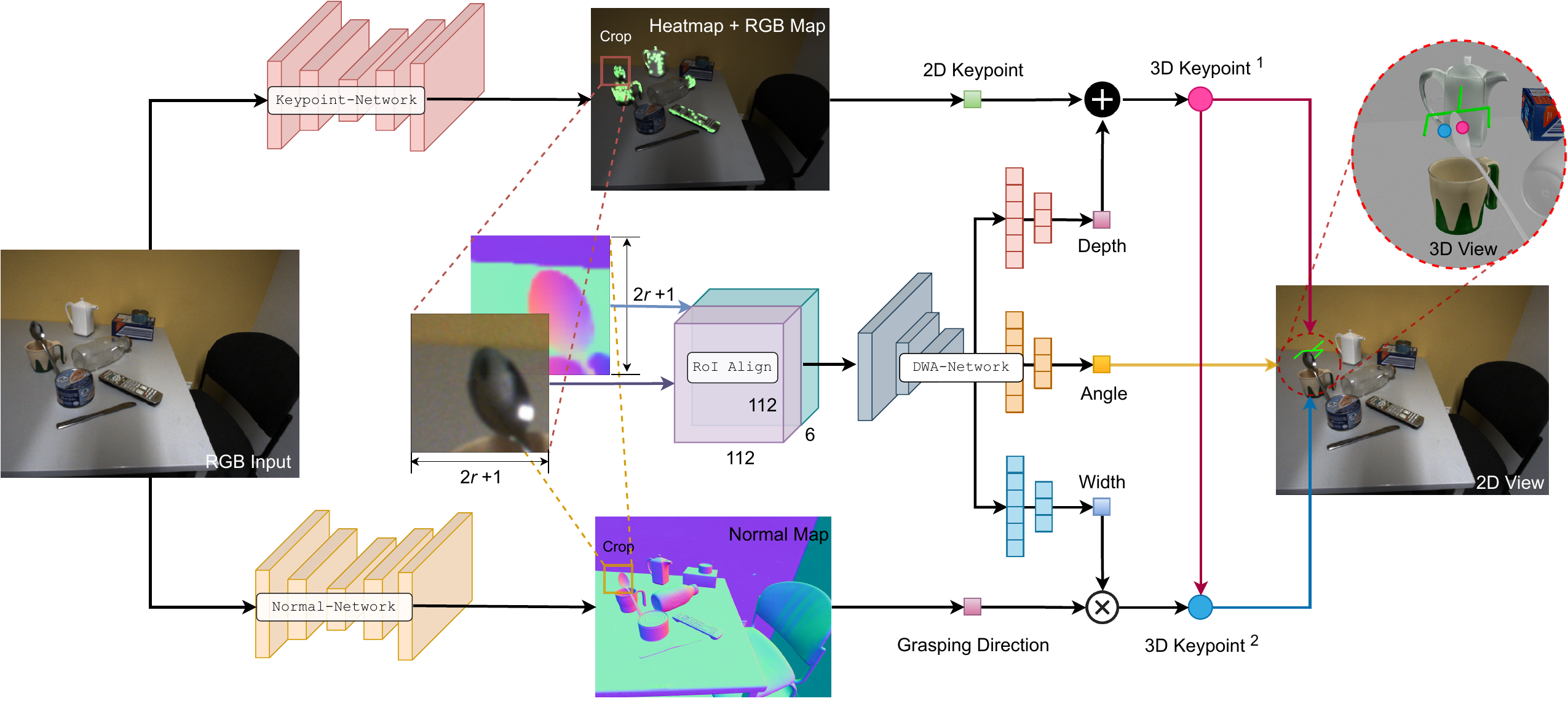}
    \caption{Schematic overview over our MonoGraspNet pipeline. Given a monocular image, our Keypoint-Network and Normal-Network predict a keypoint heatmap and a normal map. After 2D keypoint selection and local region cropping, the DWA-Network regresses the rest grasping instructions. Exemplary for one keypoint, given the detected keypoint location, we crop the same regions in RGB image and normal map using an adjustable radius $r$ to obtain a $2 \times (2r+1) \times (2r+1) \times 3$ feature map, which we then reshape to $(2r+1) \times (2r+1) \times 6$. Besides, we employ RoI Align to aggregate the features and bring them to a size of $112\times112\times6$, the same as for the other crops. The DWA-Network utilizes three branches for regressing depth, width, and angle associated with the estimated keypoint. Finally, the visible grasping point (3D Keypoint$^1$) and invisible grasping point (3D Keypoint$^2$) can be derived.}
    \label{fig:data_pipeline}
    \vspace{-3mm}
\end{figure*}

In this paper, we propose \emph{MonoGraspNet}, the first deep learning pipeline for 6-DoF grasping that requires only an RGB image to estimate accurate grasp poses. MonoGraspNet stacks two parallel networks to respectively predict 2D keypoints and surface normals. Then, it adopts a regression network to estimate keypoint depths, grasping directions, widths, and angles. The sparse setup further encourages accurate grasping property estimation on the cropped regions of interest, as shown in Fig.~\ref{fig:data_pipeline}. Our method is thus more economical in training and more lightweight in deployment than those working on dense per-pixel estimation.

Furthermore, despite existing datasets~\cite{mahler2017dex, levine2018learning, eppner2021acronym, depierre2018jacquard, WinNT, zhai20222, sajjan2020clear, fang2022transcg}, we notice an absence of objects of varying photometric challenges and perfect annotation labels, preventing current methods from being used for truly arbitrary object grasping. 
Most importantly, previous datasets are not applicable to high-level tasks, like Robot-Environment Interaction, due to their limited perceptual views and single-scenario configurations.
To serve both basic and advanced manipulation tasks, we provide multi-view and multi-scene annotations with large-scale grasping labels based on two public indoor datasets~\cite{wang2022phocal, jung2022my}, which comprise challenging objects of different photometric complexity. 
Along with other off-the-shelf dense annotations, the extended dataset can serve better for diverse grasping purposes.  

In summary, our main contributions are the followings:
\begin{itemize}
    \item We propose the first deep learning pipeline for 6-DoF grasping from a single RGB image, designing a novel grasp representation.
    \item We overcome grasping limitations for photometrically challenging objects (see Fig.~\ref{fig:cover} for a first impression). Our experiments show that MonoGraspNet can achieve competitive results in grasping common objects and surpass the depth-based competitor by a large margin in grasping photometrically challenging objects.
    \item We open-source a large-scale grasping dataset comprising objects of varying photometric complexity. The data includes approximately 20M grasping labels for 44 household sequence collections under multiple views\footnote{{\url{https://sites.google.com/view/monograsp/dataset}}}.
\end{itemize}

\section{Related Work}
The purpose of arbitrary/universal grasping is to extract shared representations for all objects in different shapes and categories without prior knowledge. Methods belonging to this field can be roughly divided by the sensor modality and grasp dimensions they use.

\noindent {\bf RGB-based 3-DoF Grasping:}
Previous methods using RGB as the input source are called planar grasping. They usually set RGB cameras in a top-down view and treat pose regression as a detection problem~\cite{7989191, kumra2017robotic, asif2018graspnet, redmon2015real, zhou2018fully}. They formulate grasping points in fixed-height oriented rectangles whose width defines the grasping width. Thus, one can use robust and developed 2D detection networks to solve the problem from where oriented rectangles can also be transferred into other formats. GKNet~\cite{xu2022gknet} regards one rectangle as three keypoints, which can then be regressed by keypoint-related networks such as CenterNet~\cite{duan2019centernet} and CornerNet~\cite{law2018cornernet}. They do not introduce depth into the pipeline, so their grasp representation is limited to 3-DoF. In contrast, our method can recover grasping poses in $SE(3)$, although we only require RGB input.

\noindent {\bf Depth-based 3-DoF Grasping:} Methods in this direction represented by Dex-Net 2.0~\cite{mahler2017dex} go further by directly deploying 2D-based networks, like GQ-CNN~\cite{mahler2017dex}, on depth image processing to perform the grasping task. As the representation is aligned with planar grasping, they still belong to 3-DoF grasping and share the same issue.

\noindent {\bf Depth-based 6-DoF Grasping:} 6-DoF grasping plays a more critical role in the grasping community, allowing robots to plan higher dexterous grasps. Two ways exist in the depth processing: one represented by~\cite{zhu20216} directly uses a depth image and leverages both 2D detection and accurate depth to estimate the rotated angles and gripper depth to recover 6-DoF poses. However, due to the top-down view and eye-on-hand requirement, some constraints from the pose representation persist, making it unsuitable in practice for some positions. Consistent with planar grasping, it is enough for the bin-picking task, but the performance is limited for more complex manipulations. The other way which has surged to solve the problem is based on point cloud processing represented by a large body of method~\cite{fang2020graspnet, breyer2020volumetric, sundermeyer2021contact, liang2019pointnetgpd, 9830843} using 3D backbones~\cite{qi2017pointnet, qi2017pointnetplusplus, alliegro2021denoise}. They aim to achieve so-called ``AnyGrasp" performance. The success rate of these approaches is high when the depth is reliable. Due to the strong dependency on the depth sensor, the performance drops severely for photometrically challenging objects.

\noindent {\bf RGB-D based 6-DoF Grasping:} Fusing RGB with depth in a completion setup~\cite{jung2021wild} is used to deal with extreme situations and photometrically challenging objects. ClearGrasp~\cite{sajjan2020clear} masks out the transparent objects in the scene using RGB features and uses estimated normals and edges to reconstruct missing depth. TransCG~\cite{fang2022transcg} collects depth using a 3D scanner and IR system, which is used to supervise the network taking fused features of inaccurate depth and RGB image. The improved performance lays the fact that RGB can alleviate the depth-missing problem. Our work takes the idea to another level, where we explore the ability of RGB on 6-DoF grasping without capturing the whole depth map.

\vspace{-0.5mm}

 \begin{figure}[t]
    \centering
    \includegraphics[width=0.48\textwidth, angle=0]{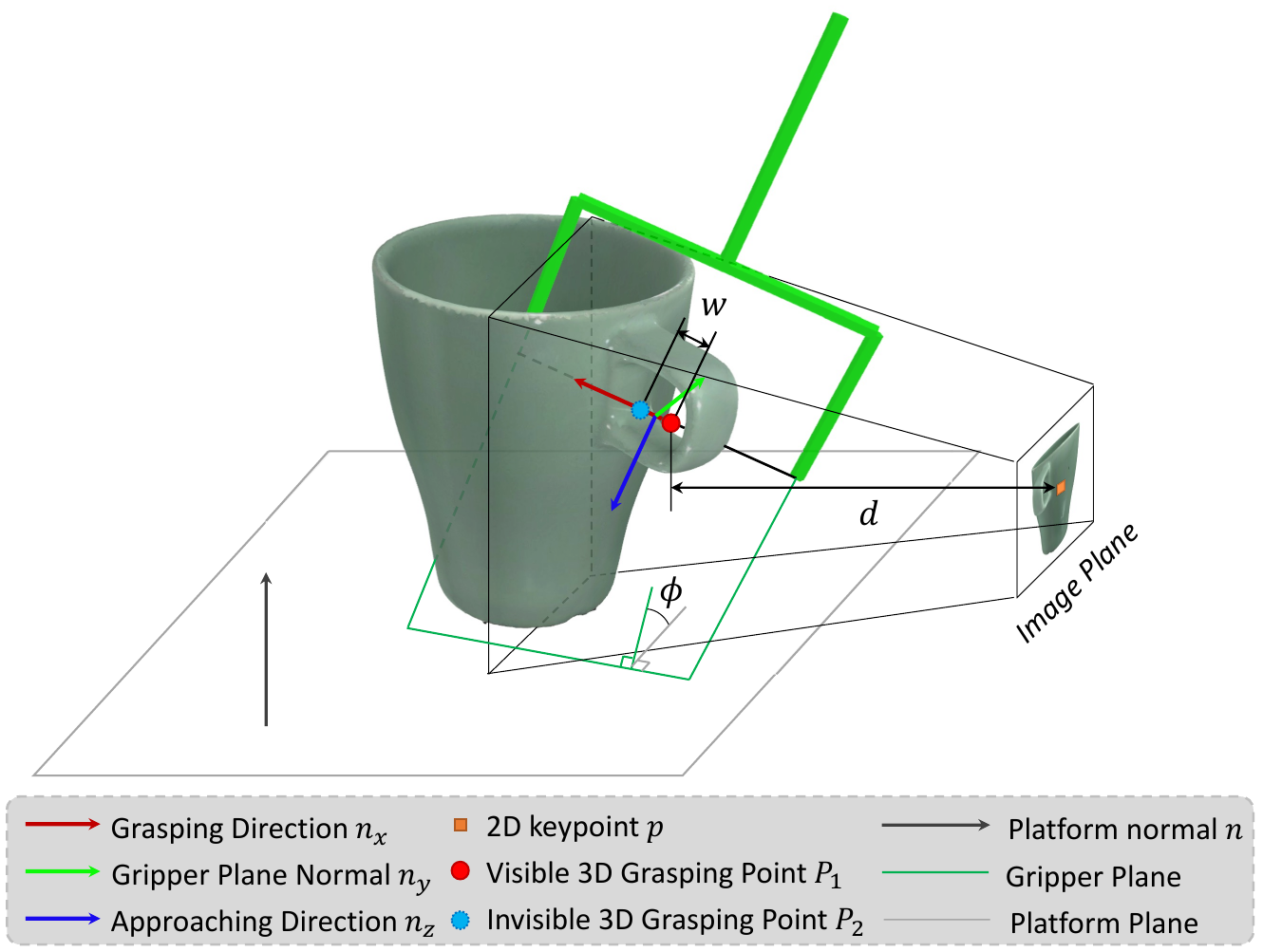}
    \caption{Our grasp representation. $d$ is the depth value of 2D keypoint $p$. $\phi$ is the dihedral angle of gripper plane (green) and platform plane (grey). $w$ is the grasping width calculated as $w=\|P_{1}-P_{2}\|$.}
    \label{fig:grasp_representation}
    \vspace{-3mm}
\end{figure}

\section{Methodology}
In this work, we propose a novel method, which we dub MonoGraspNet, for estimating the 6-DoF grasp poses from a single monocular image, and a new dataset with multi-view, multi-scene, and large-scale grasping labels.

 
\subsection{MonoGraspNet}\label{MonoGraspNet}
Given an RGB image, our method estimates a set of 6-DoF grasping representations, which are then used to recover 6-DoF poses leveraging a simple conversion. The input image is fed to the Keypoint- and Normal-Network to estimate a set of keypoints represented by a heatmap and a normal map. Those keypoints are then used to crop RGB and normal maps into joint patches, which are used as the input to our DWA-Network for estimating depth, width, and angle (DWA). The system pipeline is illustrated in Figure~\ref{fig:data_pipeline}.

\noindent
\textbf{Grasp Representation:} To better suit our setup, we introduce a novel representation, combining and extending the different representations from Contact-GraspNet~\cite{sundermeyer2021contact} and L2G~\cite{9830843} to derive our new formulation (TABLE~\ref{tab:repre}).
Contact-GraspNet uses the visible grasping point $P_1$, grasping axis $\mathbf{n_x}$, approaching axis $\mathbf{n_z}$ and grasping width $w$ to represent the 6-DoF grasp, whereas L2G relies on two 3D grasping keypoints $P_1, P_2$ together with the angle between gripper plane and platform plane $\phi$. As recovering sensitive 3D information from 2D input is a difficult ill-posed problem, the pattern we propose in this work is to leverage 2D features to a large extent. Specifically, we split $P_1$ into a 2d keypoint $p$ aligned with planar grasping represented by~\cite{xu2022gknet} and a further depth value $d$, ending up with $\{\mathbf{n_x}, p, d, w , \phi\}$ as our final parameterization, as shown in Fig.~\ref{fig:grasp_representation} (see Sec.~\ref{Recovery}for 6-DoF pose recovery).

\begin{table}[t]
\centering
\caption{Grasp Representation Comparison with others}
\small
\begin{tabular}{@{}llcc@{}} \toprule
  Methods  & Input & \#Format \\ \midrule
    C-GraspNet~\cite{sundermeyer2021contact} & Point Cloud  & $P_1, \mathbf{n_x}, \mathbf{n_z}, w$ \\ 
    L2G~\cite{9830843} & Point Cloud  & $P_1, P_2, \phi$ \\
 &&& \\
    MonoGraspNet (Ours) & RGB  & $\mathbf{n_x}, p, d, w , \phi$ \\
     \bottomrule
\end{tabular}
\label{tab:repre}
\end{table}

\noindent
\textbf{Direction Regression: }Robust regression of the grasping axis $\mathbf{n_x}$, which describes the gripper closure direction, is an important component in obtaining our grasp pose. It can be calculated by antipodal sampling through force closure inspection, assuming the availability of an object mesh. For parallel-jaw grippers, force closure depends on the friction cone and the direction of the line connecting two grasping points. Furthermore, the friction cone is determined by the object surface normal $\mathbf{v}\in \mathbb{R}^{3}$ and the friction coefficient $\mu$, which are unknown in the real-world implementation. Nonetheless, the smaller the cosine distance between $\mathbf{n_x}$ and $-\mathbf{v}$, the higher the grasping success rate, regardless of $\mu$. Thus, we can transform the problem of direction regression to the problem of normal estimation, treating $\mathbf{n_x} =  -\mathbf{v^\ast} \approx -\mathbf{v}$, with $\mathbf{v^\ast}$ being the estimated surface normal. We employ our normal estimation network (Normal-Network), shown in Fig.~\ref{fig:data_pipeline}, to infer the normal map. Since the normal consistency around the edges tends to be lower than other parts, grasping can become very complicated for the robot. To alleviate this problem, we use detected 2D keypoints to search for the corresponding normal, as discussed below.

\noindent
\textbf{Keypoint Detection: }Successful grasping is highly dependent on the accurate prediction of the underlying 3D geometry. Unfortunately, leveraging a state-of-the-art depth estimator is not a promising direction due to the bleeding-out effect around depth discontinuities. It is proved in Sec.~\ref{exper}.
Therefore, we instead focus on the utilization of 2D features. In particular, we propose to detect 2D keypoints $p$ first and subsequently recover the associated visible 3D keypoints $P_1$, as shown in Sec.~\ref{Recovery}. Notice that we formulate keypoint detection as heatmap regression using our Keypoint-Network, as depicted in Fig.~\ref{fig:data_pipeline}.
Thanks to dense and accurate annotations in our dataset, we observe that ground truth keypoints spread on the smooth surfaces of objects instead of sharp edges. By using estimated keypoints to query surface normals after properly training the network and inspecting the normal consistency around these keypoints, we can overcome the issue mentioned above in the part of normal vector selection.

\noindent
\textbf{DWA Regression: }In order to obtain the final pose, after having estimated the normal, we extra need to predict depth, width, and angle (DWA). NeWCRFs~\cite{yuan2022newcrfs}, a Conditional Random Field (CRF) based method, has shown that: (i) the value to be estimated coming from RGB relies on cues such as colors and pixel positions. Further, (ii) the depth of a pixel is usually not determined by distant pixels but by pixels that are only within a certain distance of the target pixel. For a single pixel, too many pairwise connections result in similar performance but require a lot of redundant computations. Based on (i), we additionally improve the idea by adding learned surface normal into consideration:
\begin{equation}
\begin{aligned}
& E(\mathbf{q}) =\sum_i E_u\left(q_i\right)+\sum_{i j} E_p\left(q_i, q_j\right) \\
& E_p =f\left(q_i, q_j\right) g\left(I_i, I_j\right) h\left(p_i, p_j\right) k\left(\mathbf{v_i}, \mathbf{v_j}\right),
\end{aligned}
\end{equation}
where $E(\mathbf{q})$ is the energy function for the pixels in the image, $q_i$ is the feature value of pixel $i$, and $j$ represents the other pixels. The unary potential function $E_u$ uses features of pixel $i$ to calculate its associated energy. Further, the pairwise potential function $E_p$ computes the energy for pairs of pixels, and $f\left(q_i, q_j\right)$ is the pairwise weight. Finally, $I_i$ denotes the RGB value of pixel $i$. $p_i$ is the coordinate of pixel $i$ and $\mathbf{v_i}$ is the normal vector at $p_i$. 
The motivation is that surface normals can provide strong prior information about objects' shapes. For extracting features from photometrically challenging objects, especially transparent ones, color and position are insufficient, as background and transparent parts can look similar in RGB. However, their orientations represented by the surface normals are different. Introducing the normal penalization $k\left(\mathbf{v_i}, \mathbf{v_j}\right)$ into the function can make original $f\left(q_i, q_j\right)$ in~\cite{yuan2022newcrfs} reweighted and optimized. 

Inspired by (ii), we crop local regions centered on each keypoint to let the feature encoder only focus on the respective regions, as our task only cares about depth estimation for part of the map. Besides, we further extend (ii) to the grasping width and angle regression. In practice, we achieve this by concatenating RGB-normalized crops and normal crops along the feature channel. $f\left(q_i, q_j\right)$ can be represented by various existed feature encoder, such as~\cite{he2016deep, dosovitskiy2020image, liu2021swin}. Before sending feature crops into the encoder, we perform RoI Align~\cite{he2017mask} on each crop, as some of the keypoints may appear near the boundaries of RGB images, resulting in these crops with different sizes. Moreover, RoI Align allows flexible feature output adjustment. After obtaining the high-dimension feature map, we employ fully connected layers to regress depth $d$, grasping width $w$ and rotated angle $\phi$ for each 2D keypoint. 

\begin{figure}[t]
    \centering
    \includegraphics[width=1.0\linewidth, angle=0]{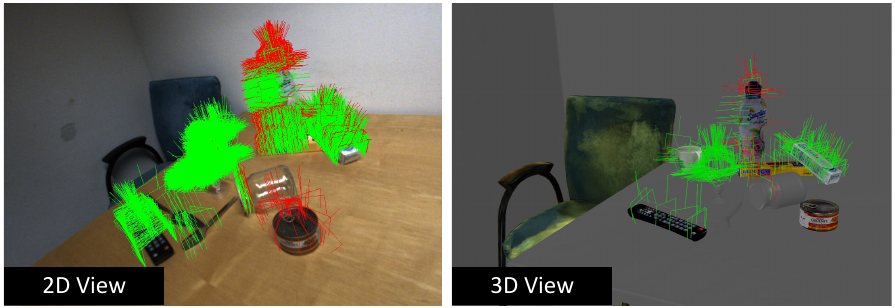}
    \caption{Examples from one scene in the proposed dataset. Failed grasps are in red, while successful grasps are in green. The first picture shows the dense annotations in the image. The second one demonstrates downsampled 6-DoF grasps in the 3D view for a better look.}
    \label{fig:dataset}
    \vspace{-3mm}
\end{figure}%

\noindent
\textbf{Network Selections: } In this work, we leverage the state-of-the-art approach~\cite{Bae2021} for normal prediction and adopt the same loss functions originally proposed, \emph{NLL loss}. For the Keypoint-Network, we use the center-point regression branch in GKNet~\cite{xu2022gknet} with DLA~\cite{yu2018deep} as the backbone. We use a variant of the \emph{focal loss} as in~\cite{law2018cornernet}. Finally, we employ an attention-based backbone Swin-Transformer Base~\cite{liu2021swin} as the backbone for our DWA-Network, as it has been proven in several regression works to be capable of acting as a very strong feature extractor. To be consistent with Swin-Transformer, we set the input after RoI Align as $112 \times 112 \times 6$ and utilize a patch size of $2$. The output feature map is of dimension $7\times 7 \times 1024$, consistent with the original paper. We use \textit{L2 loss} for all three branches in our DWA-Network.

\subsection{6-DoF Pose Recovery}
\label{Recovery}
The visible and invisible 3D keypoints $P_1$ and $P_2$ can be obtained by
\begin{equation}
P_1 = d K^{-1} p, \quad P_2 = P_1 - w \mathbf{v}^\ast,
\end{equation}
where $K$ is the camera intrinsic matrix. Finally, using $P_1$, $P_2$, and $\phi$, we can now calculate the 6-DoF grasp pose. Note that $\phi$ is collected in the robot base frame, with $P_1$ and $P_2$ in the camera frame. So we first transform $P_1$ and $P_2$ to robot base and calculate the center point $P_{c}^{\ast}$ according to
\begin{equation}
\begin{aligned}
P_{c}^{\ast} &= (P_{1}^{\ast} + P_{2}^{\ast}) / 2 \\
[P_{1}^{\ast},  P_{2}^{\ast}] &= \mathbf{T}_{ base \leftarrow { cam}} [P_1, P_2].
\end{aligned}
\end{equation}

We can also obtain each component vector in the rotation matrix for the grasp by:
\begin{equation}
\begin{aligned}
&\mathbf{n_x}=-\mathbf{a^*},
\end{aligned}
\end{equation}
\begin{equation}
\begin{aligned}
&\left\{\begin{array}{l}
\mathbf{n_y} \cdot \mathbf{n_x}=0 \\
\phi =\arccos{(\mathbf{n_y} \cdot \mathbf{n})}\\
\left\|\mathbf{n_y}\right\|=1
\end{array}\right. ,\\
\end{aligned}
\label{ny}
\end{equation}
\begin{equation}
\begin{aligned}
&\mathbf{n_z}=\mathbf{n_x} \times \mathbf{n_y},
\end{aligned}
\end{equation}
where $\mathbf{n}$ is the normal of the platform plane $[0,0,1]^\text{T}$. From~\eqref{ny}, two $\mathbf{n_y}$ solutions are obtained by solving a quadratic equation. We always take the one whose accompanying $\mathbf{n_z}$ points to the platform since it is safer and easier for the robot to reach. Then the grasp rotation matrix can be represented by $\mathbf{R} = [\mathbf{n_x}, \mathbf{n_y}, \mathbf{n_z}]$. The final 6-DoF pose $\mathbf{G} = {\left[ {{\mathbf R},P_{c}^{\ast};{\mathbf 0},1} \right]}$ in the robot base frame.

\subsection{Dataset Collection}

To explore RGB-only information in this challenging grasping task, we need to learn 3D information from a dataset that contains pixel-perfect depth rather than the measurement coming from a regular depth camera, as it does not work when facing photometrically challenging objects. Moreover, the data should preferably originate from the real world to avoid creating any synthetic-to-real domain gap.

Given this, we choose PhoCaL~\cite{wang2022phocal} and HAMMER~\cite{jung2022my} as our base dataset, which contains various accurate geometry data. 
Meanwhile, As a robot arm collected both datasets in real-world scenarios, the perception field and manipulation workspace are consistent with the grasping task. By taking advantage of this, we label grasping poses for each mesh in the dataset by antipodal sampling and inspect poses with physical simulators like~\cite{makoviychuk2isaac,miller2004graspit}, which is a well-developed routine in~\cite{eppner2021acronym, eppner2019billion}. Then we reproject these grasps back into the scene in the robot frame based on 6-DoF object poses $\mathbf{T}_{ {base } \leftarrow obj} $, and perform collision inspection with background meshes. After these procedures, we can obtain grasp labels in each camera frame $\mathbf{G}_{cam}$ by recorded camera poses relative to the robot base $\mathbf{T}_{cam \leftarrow { base }}$: 

\begin{equation}
\begin{aligned}
\mathbf{G}_{cam} &=\mathbf{T}_{cam \leftarrow obj} \mathbf{G}_{obj}\\
\mathbf{T}_{cam \leftarrow obj} &=\mathbf{T}_{cam \leftarrow { base }} \mathbf{T}_{ {base } \leftarrow obj}.
\end{aligned}
\end{equation}
An example is shown in Fig.~\ref{fig:dataset}. This extended dataset will be available with original~\cite{wang2022phocal} and~\cite{jung2022my} for the community to research.

As described in Sec.~\ref{MonoGraspNet}, our pipeline calculates grasps by using visible 2D features from which we extract invisible information. To serve this purpose better, We further search for the nearest 3D points with respect to the grasping points in the point cloud, which can be obtained from depth ground truth with intrinsic camera parameters, and then reproject these points to the image plane to get corresponding pixels. These pixels are treated as the ground truth of visible 2D keypoints. Moreover, we record the correspondence between each keypoint and the corresponding grasping point pair to attain the ground truth of grasping width and grasping angle for each grasp.

\section{Experimental Evaluation}
\label{exper}
In this section, we first provide our utilized implementation details. Then we introduce our experimental setup and present our results compared with the depth-based state-of-the-art grasping method Contact-GraspNet in various real-world experiments, defined as the ultimate test of the grasp performance in~\cite{mousavian20196}.

\subsection{Implementation Details}

\noindent
\textbf{Training Settings:} 
All the experiments are conducted on a single NVIDIA RTX 3090 GPU with Adam optimizer.
The Keypoint-Network is trained for 30 epochs with an initial learning rate of 1e-4 and batch size of $2$.
For Normal-Network training, we follow the instructions in the original paper~\cite{Bae2021}.
The proposed DWA-Network is trained for 30 epochs with an initial learning rate to be 1.25e-4 and batch size to be $32$.
We decrease the learning rate by $10$ times at the $15$th and the $20$th epoch. 
Our input image size is $832 \times 1088$. 

\noindent
\textbf{Robot Hardware:} We use a 7-DoF Franka Panda robot with a parallel-jaw gripper as the end-effector. The RGB camera mounted on the gripper base is a Phoenix 5.0 MP Polarization camera, the same as the one used in~\cite{wang2022phocal, jung2022my} shown in Fig.~\ref{fig:cover}. For testing Contact-GraspNet, we install an Azure Kinect depth camera beside the robot. All cameras are hand-eye calibrated.

\begin{table*}[ht]
\centering
\caption{Success rate (\%) of familiar single object grasping experiments in seen and unseen scenes}
\begin{tabular}{cccccccccccccccccccc}
\hlineB{2}
\multirow{2}{*}{Method}  &  \multicolumn{2}{c}{\textbf{Avg.}} &  \multicolumn{2}{c}{ (1) Cocktail glass} &  \multicolumn{2}{c}{(2) Mug} &  \multicolumn{2}{c}{(3) Liquid} &  \multicolumn{2}{c}{(4) Spoon} &  \multicolumn{2}{c}{(5) Monster} &  \multicolumn{2}{c}{(6) Biscuit} \\ \cline{2-15}
& S & Un-S & S & Un-S & S & Un-S & S & Un-S & S & Un-S & S & Un-S & S & Un-S   \\ \cline{1-15}
\cite{yuan2022newcrfs}+C-GraspNet~\cite{sundermeyer2021contact} &  23.3 & 10.0 & 0 & 0 & 0 & 13.3 & 33.3 & 26.7 & 6.7 & 0 & 26.7 & 0 & 73.3 & 20.0\\
C-GraspNet~\cite{sundermeyer2021contact} &  62.2 & 60.0 & 6.7 & 0 & \textbf{93.3} &  \textbf{93.3}  & \textbf{60.0} & \textbf{66.7} & 20.0 & 6.7 & \textbf{93.3} & \textbf{93.3} & \textbf{100} & \textbf{100} \\
MonoGraspNet (Ours) & \textbf{83.9} & \textbf{72.2} & \textbf{80.0} & \textbf{60.0} & 86.7 & 86.7 & \textbf{60.0} & 53.3 & \textbf{86.7} & \textbf{80.0}  & 86.7 & 73.3 & 93.3 & 80.0\\ \hlineB{2}
\end{tabular}
\label{tab:single_grasp1}
\vspace{-1mm}
\end{table*}
\vspace{-0.5mm}

\begin{table*}[ht]
\centering
\caption{Success rate (\%) of unfamiliar single object grasping experiments in seen and unseen scenes}
\begin{tabular}{cccccccccccccccccccc}
\hlineB{2}
\multirow{2}{*}{Method} &  \multicolumn{2}{c}{\textbf{Avg.}} &  \multicolumn{2}{c}{(8) Narrow glass} &  \multicolumn{2}{c}{(9) Red kettle} &  \multicolumn{2}{c}{(10) Trans-kettle} &  \multicolumn{2}{c}{(11) Glass} &  \multicolumn{2}{c}{(12) Fanta} &  \multicolumn{2}{c}{(13) Gel}\\ \cline{2-15}
& S & Un-S & S & Un-S & S & Un-S & S & Un-S & S & Un-S & S & Un-S & S & Un-S  \\  \cline{1-15}
\cite{yuan2022newcrfs}+C-GraspNet~\cite{sundermeyer2021contact} &  26.2 & 11.9 & 0 & 0 & 30.7 & 13.3 & 0 & 0 & 13.3 & 6.7 & 40.0 & 20.0 & 73.3 & 40.0\\ 
C-GraspNet~\cite{sundermeyer2021contact} &  42.2 & 42.2 & 0 & 0 & \textbf{86.7} & 80.0 & 6.7 & 0 & 0 & 0 & \textbf{93.3} & \textbf{100} & 66.7 & 73.3\\
MonoGraspNet (Ours) & \textbf{75.5} & \textbf{72.2} & \textbf{60.0} & \textbf{40.0} & 80.0 & \textbf{86.7} & \textbf{60.0} & \textbf{53.3} & \textbf{80.0} & \textbf{80.0} & 86.7 & 93.3 & \textbf{80.0} & \textbf{80.0}\\ \hlineB{2}
\end{tabular}
\label{tab:single_grasp2}
\end{table*}

\begin{table}[t]
\centering
\caption{Completion rate of clutter removal experiments}
\begin{tabular}{cccccccc}
\hlineB{2}
\multirow{2}{*}{Type} & \multicolumn{2}{c}{C-GraspNet~\cite{sundermeyer2021contact}}  & \multicolumn{2}{c}{MonoGraspNet (Ours)} \\ \cline{2-5} 
& 1st trial & 2nd trial & 1st trial  & 2nd trial  \\ \hline
\multicolumn{1}{c}{4-object scene} & 3/4     & 3/4       & 3/4     & 3/4    \\
\multicolumn{1}{c}{5-object scene} & 4/5    & 4/5       & 4/5   & \textbf{5/5}    \\ \hlineB{2}
\end{tabular}
\label{tab:clutter_grasp}
\vspace{-3mm}
\end{table}

 \begin{figure}[t]
    \centering
    \includegraphics[width=\linewidth, angle=0]{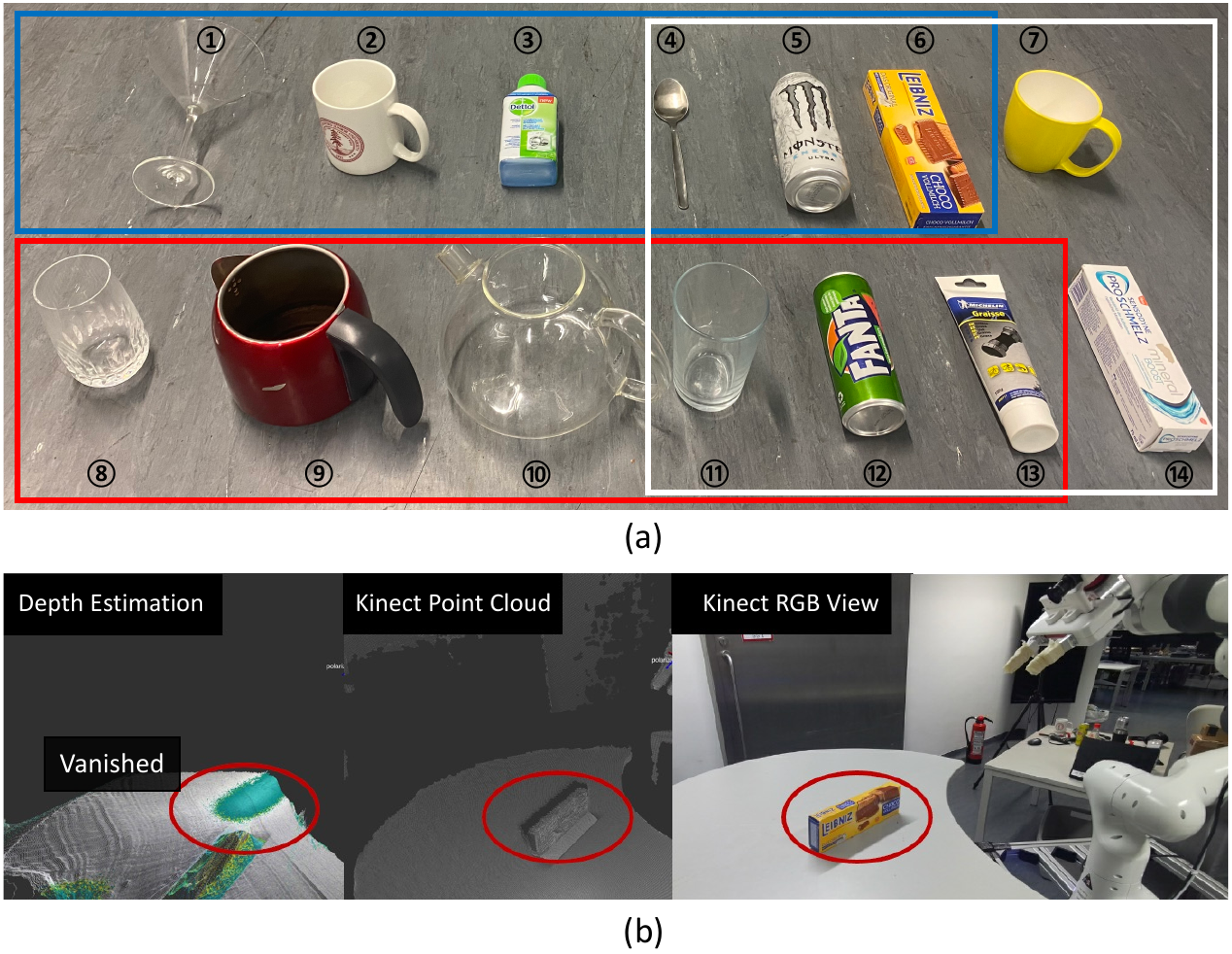}
    \caption{a) Objects used for testing. 1-6 are familiar objects (blue block). 8-13 are unfamiliar objects (red block). Objects in white block are used for clutter removal. b) Recovered point cloud from monocular depth estimation is out of the shape, resulting in the biscuit box vanished from the scene.}
    \label{fig:some}
    \vspace{-3mm}
\end{figure}

\subsection{Experimental Details}
\noindent
\textbf{Evaluation Protocols:}
We select 14 objects shown in Fig.~\ref{fig:some}.a) to compare our method with Contact-GraspNet and report the \emph{success rate} and \emph{completion rate}. The success rate is used for testing single-object grasping in seen and unseen scenarios. We let the robot execute 15 grasps by putting the objects at three random positions in the scene, and then we calculate successful grasps. The completion rate is the percentage of objects removed from the clutter, which can be a robust metric when testing multi-object grasping. We execute the grasping task in four scenarios with different configurations sorted by ascending difficulty:  1) familiar objects in seen scenes, 2) familiar objects in unseen scenes, 3) unfamiliar objects in seen scenes, and 4) unfamiliar objects in unseen scenes. We additionally set an experiment by combining Contact-GraspNet and NeWCRFs~\cite{yuan2022newcrfs} to support our claim in Sec.~\ref{MonoGraspNet}-Keypoint Detection, which is that the path of estimating dense depth map and grasping objects using the recovered point cloud results in bad performance.

\noindent
\textbf{Running Pipeline:}
In our experiment, we directly use basic joint control instead of advanced options that require 3D information for the motion planning part. To ensure impartiality, we apply the same planning method for Contact-GraspNet.
Since there is no segmentation module inside MonoGraspNet, we also remove the relevant module~\cite{xiang2021learning} accompanying Contact-GraspNet, which is allowed by the instruction from~\cite{sundermeyer2021contact}.

\noindent
\textbf{Results} 
TABLE~\ref{tab:single_grasp1} and TABLE~\ref{tab:single_grasp2} show the comparison results under the four situations.
\textit{Familiar objects:} trained objects, but in untrained views.
\textit{Unfamiliar objects:} untrained objects.
\textit{S:} scenes that were seen in the training set but in a novel view.
\textit{Un-S:} unseen scenes.
As shown in Row 1, directly feeding the point cloud from~\cite{yuan2022newcrfs} into Contact-GraspNet yields the worst results, with only $23.3\%$ success rate for familiar objects in seen scenes, as the recovered point cloud is distorted in an unstructured shape. An example is shown in Fig.~\ref{fig:some}.b).
In TABLE~\ref{tab:single_grasp1}, it can be seen that Contact-GraspNet is slightly superior to MonoGraspNet when grasping normal objects, i.e., objects (2, 5, 6), with an average of $6.6\%$ leap forward. 
This seemingly discouraging outcome may be expected since depth from sensors is typically more accurate than predicted depth from RGB images.
However, when dealing with photometrically challenging objects (1, 3, 4), MonoGraspNet surpasses Contact-GraspNet by a large margin, with an average success rate of $78.9\%$  against $28.9\%$.
In this situation, inaccurate depth limits the performance of grasp pose estimation of Contact-GraspNet.

We also verified the generalization ability of MonoGraspNet.
By comparing the results in unseen scenes with the ones in seen scenes, we can see that the success rate remains stable and consistent, with only a maximum $11.7\%$ decrease in average.
It proves that MonoGraspNet can generalize to other real-world scenarios without pre-training or fine-tuning.
For unfamiliar object grasping in TABLE~\ref{tab:single_grasp2}, even though the textures or shapes of these objects are distinct from the trained ones in the dataset, MonoGraspNet can still yield an adequate success rate, as it uses local 2D features to infer grasps, which remains stable across different objects and different scenes.

We further perform a more challenging clutter removal task. As we do not have a collision check module, we always let the robot execute the nearest grasp. We randomly choose four or five objects with one challenging object inside from the white block of Fig.~\ref{fig:some}. TABLE~\ref{tab:clutter_grasp} shows that we still show a competitive performance compared to Contact-GraspNet when moving simple clutter with photometrically challenging objects inside.

\begin{table}[t]
\centering
\caption{Success rate (\%) comparison with different depth estimation}
\begin{tabular}{cccccc}
\hlineB{2}
\multirow{2}{*}{Method} & \multicolumn{2}{c}{Normal}  & \multicolumn{2}{c}{Challenging} \\ \cline{2-5} 
\multirow{2}{*}{} & Fanta & Biscuit & Spoon & Glass\\ \hline
\multicolumn{1}{c}{DWA-Net w/o normal} & 80.0     & 93.3           & 86.7     & 66.7           \\
\multicolumn{1}{c}{DWA-Net} & \textbf{86.7}     & 93.3         & 86.7   &   \textbf{80.0}         \\ \hlineB{2}
\end{tabular}
\label{tab:ablation}
\vspace{-2mm}
\end{table}

\subsection{Ablation Study}
To prove the motivation that introducing the normal map can help depth estimation in DWA-Network. We re-train a DWA-Network with RGB crops as the only input. TABLE~\ref{tab:ablation} shows that the success rate performance of grasping normal objects and even the reflective object (Spoon as an example here) are similar, but the improvement is obvious for grasping transparent objects (unfamiliar Glass).

\section{Conclusions}

In contrast to previous conventional routines which rely on accurate depth, this paper proposes the first RGB-only 6-DoF grasping pipeline called MonoGraspNet. It utilizes stable 2D features to simultaneously handle arbitrary object grasping and overcome the problems induced by photometrically challenging objects. 
Given an RGB image, our method estimates a keypoint map and a normal map to recover the 6-DoF grasping poses represented by our novel representation.
The experiments demonstrate that our method can achieve competitive results on grasping common objects with simple textures while surpassing the depth-based competitor significantly on grasping photometrically challenging objects.
We additionally annotate a large-scale real-world grasping dataset, containing 44 object settings of mixed photometric complexity with approximately 20M accurate grasping labels.



\bibliographystyle{IEEEtran}
\bibliography{IEEEabrv,references}

\end{document}